%% file: main.tex
\documentclass{article} 
\usepackage{iclr2025_conference,times}

\input{math_commands.tex}

\usepackage{hyperref}
\usepackage{url}

\usepackage{booktabs}
\usepackage{multirow}
\usepackage{graphicx}
\usepackage{wrapfig}
\usepackage{caption}
\usepackage{subcaption}

\newcommand{\ie}{i.e.\ }

\title{LeC$^2$O-NeRF: Learning Continuous and \\ Compact Large-Scale Occupancy for \\ Urban Scenes}

\author{Zhenxing Mi \& Dan Xu \\
Department of Computer Science and Engineering \\ 
The Hong Kong University of Science and Technology (HKUST)\\
Clear Water Bay, Kowloon, Hong Kong \\
\texttt{zmiaa@connect.ust.hk, danxu@cse.ust.hk}
}
\iclrfinalcopy 
\begin{document}

\maketitle

\begin{abstract}

In Neural Radiance Fields (NeRFs), a critical problem is how to effectively estimate the occupancy to guide empty-space skipping and point sampling. The grid-based methods work well for small-scale scenes. However, on large-scale scenes, they are limited by predefined bounding boxes, grid resolutions, and high memory usage for grid updates, and thus struggle to speed up training for large-scale, irregularly bounded and complex urban scenes without sacrificing accuracy. In this paper, we propose to learn a continuous and compact large-scale occupancy network, which can classify 3D points as occupied or unoccupied points. We successfully train this occupancy network end-to-end together with the radiance field in a self-supervised manner by three core designs. \emph{First}, we propose a novel imbalanced occupancy loss to regularize the occupancy network. It enables the occupancy network to effectively control the ratio of the unoccupied and occupied points, motivated by the prior that most of the 3D scene points are unoccupied. \emph{Second}, we design an imbalanced network architecture containing a large scene network and a small empty space network to separately encode occupied and unoccupied points classified by the occupancy network. This imbalanced structure can effectively model the imbalanced nature of occupied and unoccupied regions. \emph{Third}, we design an explicit density loss to guide the occupancy network, making the density of unoccupied points smaller. As far as we know, we are the first to learn a continuous and compact occupancy of large-scale NeRF by a network. We show in the experiments that our occupancy network can very quickly learn more compact, accurate and smooth occupancy compared to the occupancy grid. With our learned occupancy as guidance for empty space skipping on several challenging large-scale benchmarks, our method consistently obtains higher accuracy compared to the occupancy grid, and our method can successfully speed up state-of-the-art NeRF methods without sacrificing accuracy.
\end{abstract}

\section{Introduction}

\begin{figure}[t]

\begin{center}
   \includegraphics[width=\linewidth]{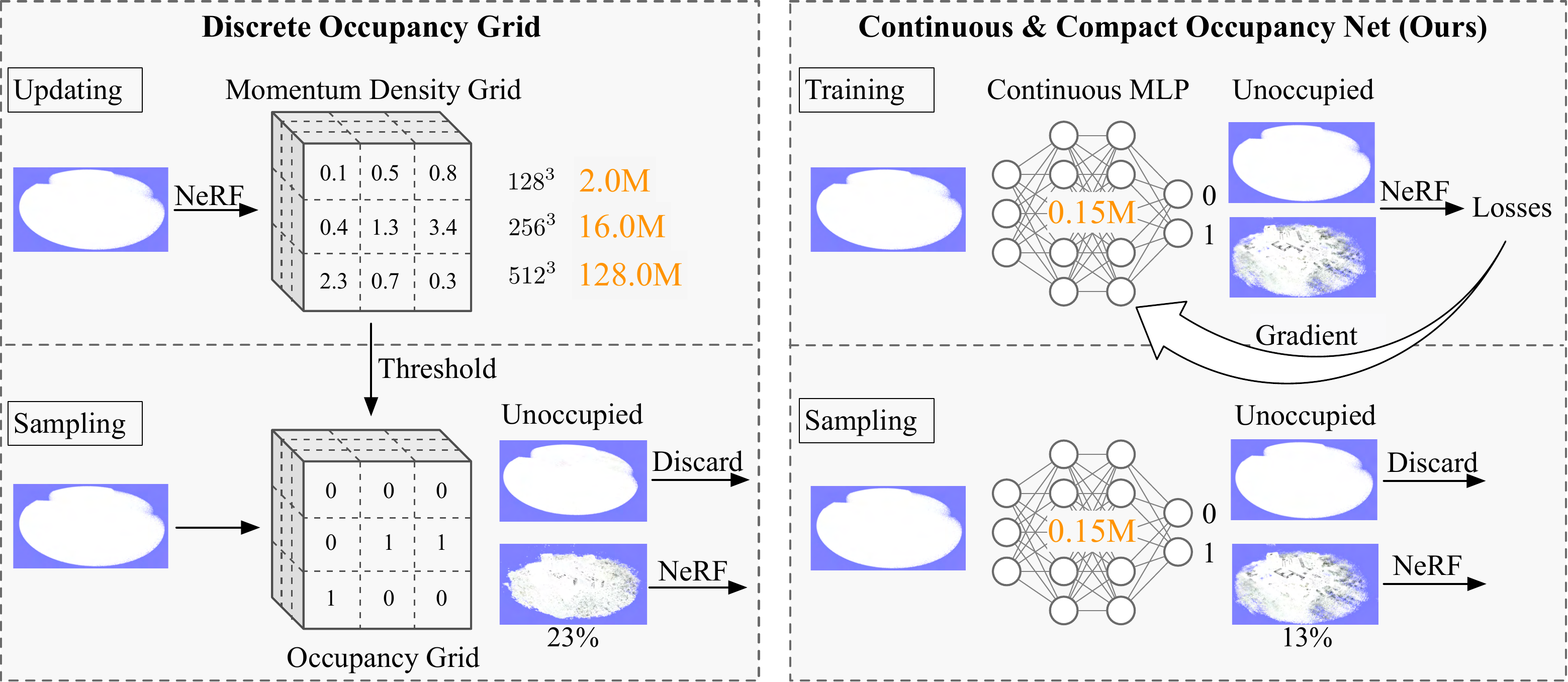}
\end{center}
   \caption{Differences between the occupancy grid~\citep{li2022nerfacc,mueller2022instant} and our occupancy network. Our occupancy network is a \emph{compact and continuous} MLP with only 0.15M parameters, trained together with NeRF networks by our designed losses. The occupancy grid is a \emph{discrete} representation and stores 2.0M and 128.0M parameters for a resolution of $128^3$ and $512^3$, respectively. It is updated by evaluating the NeRF network and is not aware of the training loss. The images are the visualization of the occupied and unoccupied points as stated in Section~\ref{sec:Metricsandvisualization}. The whole points are sent to the grid or the occupancy network and are split into two parts of occupied and unoccupied points.}
\label{fig:teaser}
\end{figure}

Neural Radiance Fields (NeRF)~\citep{mildenhall2020nerf} have been used to model large-scale 3D scenes~\citep{Turki_2022_CVPR, Tancik_2022_CVPR,mi2023switchnerf}. 
Although achieving promising performances, the critical problem of modeling occupancy for large-scale scenes remains under-explored.
A large 3D scene is usually very sparse, with a large portion of the 3D scene as empty spaces. Thus, modeling the occupancy can effectively guide the empty-space skipping and point sampling. Using an occupancy grid for guided sampling has become a common practice in small-scale NeRF~\citep{mueller2022instant,Fridovich_Keil_2022_CVPR,Hu_2022_CVPR,li2022nerfacc}.
As shown in Fig.~\ref{fig:teaser}, the occupancy grid stores momentum density and occupancy in its cells. During NeRF training, points are sampled from grid cells, and the grid's density values are updated in a momentum-based manner by evaluating the NeRF model. Binary occupancy is determined by applying a threshold to the momentum density values. The computation of updating the grid is determined by the grid's resolution, and a higher resolution leads to significantly increased overhead. 

The occupancy grid works well on small-scale scenes while having clear limitations on large-scale scenes: \textbf{(i)} The memory and computation used to store and update the grid increase remarkably along with the grid's resolution. This limits the grid from increasing its resolution to model detailed large-scale scenes. \textbf{(ii)} The occupancy grid needs more prior knowledge of the scene. The scene should be more regular so that it has a tight bounding box. \textbf{(iii)} Most of the grids are unoccupied due to the scene's sparsity, making the grid not compact enough, thus wasting memory and computation. \textbf{(iv)} The momentum updating of the occupancy grid is not aware of the rendering loss, making it agnostic to the rendering quality, leading to unsatisfactory results. Due to these limitations, the occupancy grid fails to speed up the training of large-scale NeRF without sacrificing the accuracy in our experiments. Therefore, it is challenging to directly model the occupancy of large-scale complex scenes with the occupancy grid. 

To tackle the challenges of modeling occupancy for large-scale scenes, in this paper, we propose LeC$^2$O-NeRF to learn a continuous and compact occupancy representation by a network, depicted in Fig.~\ref{fig:teaser}. An essential nature of a 3D scene is that the occupied points are much fewer than the unoccupied points, while containing significantly more important information. Therefore, modeling occupancy is naturally very imbalanced. This motivates us to propose an imbalanced occupancy loss, an imbalanced network, and a density loss to successfully learn the occupancy. 
Our contributions are discussed below.

\textbf{Firstly}, we propose to learn the occupancy by a continuous and compact classification network. We train this network end-to-end and in a self-supervised manner together with the NeRF network. 
\textbf{Secondly}, we propose an imbalanced occupancy loss to regularize the occupancy network. Since a large portion of the 3D space is unoccupied, the occupancy network should explicitly model the imbalance of occupancy. We design an imbalanced occupancy loss to approximately control the portion of occupied and unoccupied points. We can use it to make only a small portion of the 3D points classified as occupied. \textbf{Thirdly}, we design an imbalanced network architecture to model the radiance field. It contains a large scene network for occupied points and a small empty space network for unoccupied points. The occupancy network works as a dispatcher to send points into different networks. A point is seen as unoccupied if the empty space network is selected for it. The empty space network contains much fewer parameters than the scene network, modeling the prior that the unoccupied points are less informative and much easier to encode. With the imbalanced occupancy loss and the imbalanced network architecture, we find that the occupancy network can already distinguish the occupied and unoccupied points effectively. 
\textbf{Fourthly}, to better learn the occupancy of a large-scale scene, we propose a density loss to guide the training of the occupancy network. In a NeRF representation, the density of an unoccupied point is much smaller than that of an occupied point. We explicitly use this constraint to design a density loss to make the occupancy network dispatch points with small density values to the empty space network. This density loss can ensure the network predicts more accurate occupancy.

\par Our imbalanced occupancy loss and the density loss work together with the rendering loss, so that our network is more aware of the rendering quality. Our LeC$^2$O-NeRF converges very fast in learning occupancy. After training the occupancy, we can utilize it to guide the point sampling in the state-of-the-art NeRF methods, such as Instant-NGP~\citep{mueller2022instant} and the large-scale Switch-NeRF~\citep{mi2023switchnerf}. We freeze the learned occupancy network and use it as an occupancy predictor. If a point is predicted as unoccupied, it is discarded and is not processed by the main NeRF network. In our experiments, we can consistently outperform the occupancy grid in terms of accuracy, and can successfully speed up state-of-the-art NeRF methods without sacrificing accuracy.
Our method can also learn more compact, accurate and smooth occupancy compared to the occupancy grid. The smoothness is apparent as shown in the rendered videos in the supplementary.

    


\vspace{-5pt}
\section{Related Work}
\vspace{-5pt}

\noindent \textbf{NeRF.} Neural Radiance Fileds~\citep{mildenhall2020nerf} utilize a multilayer perceptron (MLP) network to encode a 3D scene from multi-view images. It has been extended to model a lot of tasks~\citep{liu2020neural,xu2022point,moconerfeccv24,kerbl3Dgaussians,zhang2022nerfusion,editingiccv2023} or even city-level large-scale scenes~\citep{Turki_2022_CVPR,Tancik_2022_CVPR,wang2024pygs, mi2023switchnerf,qu2024implicit}. The main idea of these large-scale NeRF methods is to decompose the large-scale scene into partitions and use different sub-networks to encode different parts, and then compose the sub-networks. The Mega-NeRF~\citep{Turki_2022_CVPR} and Block-NeRF~\citep{Tancik_2022_CVPR} manually decompose the scene by distance or image distribution. The sub-networks are trained separately and composed with manually defined rules. The Switch-NeRF~\citep{mi2023switchnerf} learns the scene decomposition by an MoE network and trains different experts in an end-to-end manner. There are also several methods~\citep{xu2023gridguided,zhang2023efficient,zhong2024cvt} employing the hash encoding~\citep{mueller2022instant} and plane encoding~\citep{Chan2022,Chen2022ECCV} while not decomposing the scene. In contrast to these existing works, our LeC$^2$O-NeRF method focuses on learning the occupancy of a large-scale scene. The learned occupancy can be used to accelerate large-scale NeRF methods.

\par\noindent\textbf{Occupancy and efficient sampling in NeRF.} 
Many methods are proposed to estimate the important regions. NeRF~\citep{mildenhall2020nerf} trains a coarse and fine network together for hierarchical sampling. The Mip-NeRF 360~\citep{Barron_2022_CVPR} designs a small proposal network to predict density and converts it into a sampling weight vector. Apart from these methods directly predicting the weight distributions, there are many methods ~\citep{mueller2022instant,Fridovich_Keil_2022_CVPR,iccvleveraging,Hu_2022_CVPR,li2022nerfacc} use the binary occupancy for sampling. 
The NerfAcc~\citep{li2022nerfacc} provided a plug-and-play occupancy grid module and has shown in extensive experiments that estimating occupancy can greatly accelerate the training of various NeRF methods. The Instant-NGP~\citep{mueller2022instant} uses multi-scale occupancy grids to encode the occupancy. These existing methods using occupancy grids typically focus on small-scale scene modeling. The occupancy grid faces problems on large-scale scenes, as described above. In this paper, we focus on learning a continuous and compact occupancy representation for large-scale scenes.

\section{The Proposed Method}

\subsection{Overview}
\begin{figure}[t]
\begin{center}
   \includegraphics[width=1.0\textwidth]{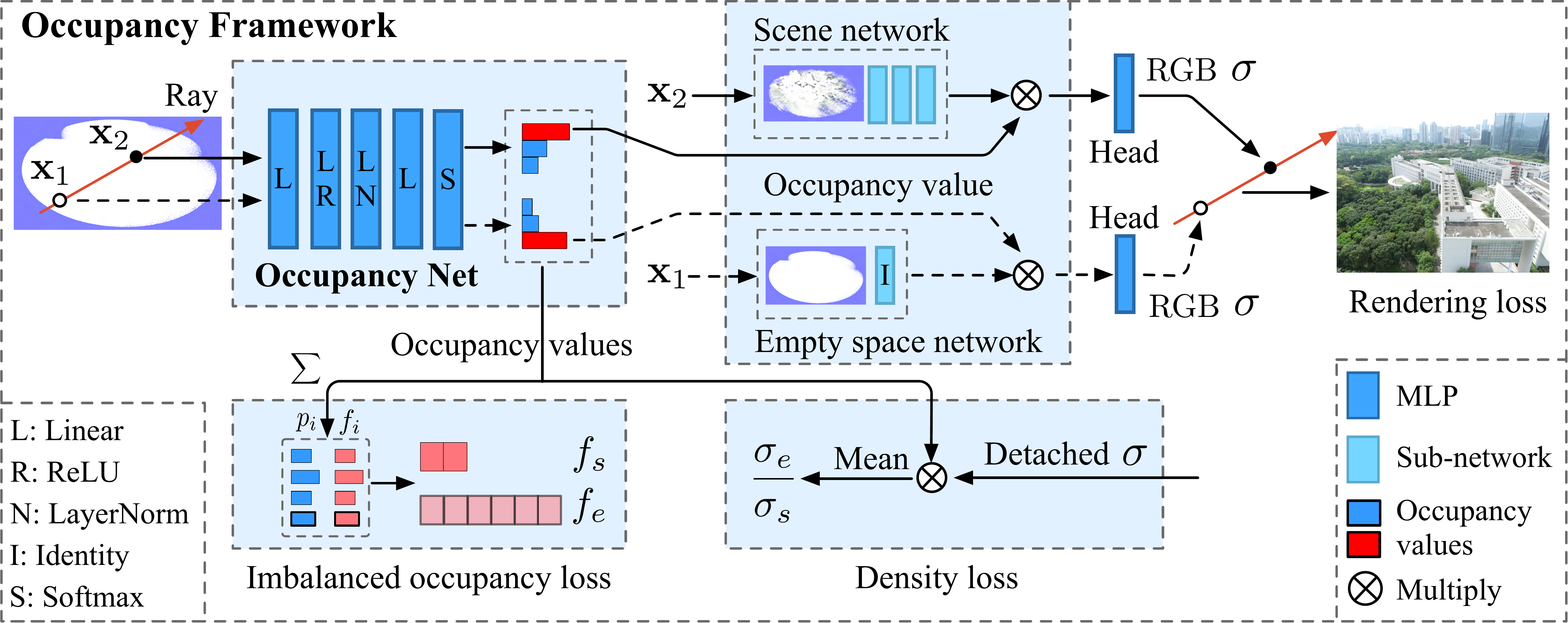}
\end{center}

   \caption{Our proposed LeC$^2$O-NeRF. The occupancy network predicts the occupancy of each point and dispatches them into different sub-networks. $\textbf{x}_1$ and $\textbf{x}_2$ go through the occupancy network and are dispatched to the empty space network and the scene network, according to their occupancy values. The occupancy can be trained end-to-end together with the NeRF network by multiplying occupancy values on the output of sub-networks. If a point is dispatched into the empty space network, it is classified as unoccupied. The occupancy network is a small MLP. We enlarge the figure of the occupancy network to clearly show its operation. The imbalanced occupancy loss and the density loss are computed by the occupancy values and the detached $\sigma$.}

\label{fig:framework}
\end{figure}

Our LeC$^2$O-NeRF learns occupancy of a 3D scene end-to-end in the training of a Neural Radiance Field $F$. The framework is shown in Fig.~\ref{fig:framework}. $F$ takes a 3D point $\textbf{x}$ and its direction $\textbf{d}$ as input. It predicts the color $\textbf{c}$ and density $\sigma$ for each $\textbf{x}$. It contains an occupancy network, $n+1$ sub-networks, and two prediction heads. The occupancy network $O$ is an MLP classification network. It dispatches different points into different sub-networks. The scene network consists of the $n$ sub-networks $\mathcal{S} = \{S_i, i=1...n\}$ and handles occupied points. The empty space network is a special tiny sub-network $E_e$ to handle unoccupied points. The prediction head $H_s$ and $H_e$ are for $\mathcal{S}$ and $E_e$, respectively. After training, the occupancy of the 3D scene is encoded in the occupancy network $O$ and the radiance field is encoded in the scene and empty space networks. 

\par A 3D point $\textbf{x}$ first goes through the occupancy network $O$ and obtains $n+1$ occupancy values. These values correspond to the scene network's $n$ sub-networks and the empty space network. Then, $\textbf{x}$ is dispatched into only one of the $n+1$ sub-networks according to the occupancy values. If a scene sub-network is selected, it implies that $\textbf{x}$ is occupied. $\textbf{x}$ is then input to the scene MLP and the prediction head $H_s$.
If the empty space network is selected, it implies that $\textbf{x}$ is not occupied. It then goes through $E_e$ and $H_e$. Therefore, the occupancy network performs as a binary classification network to identify the occupied and unoccupied points. Our proposed imbalanced occupancy loss and the density loss are computed to train the occupancy network together with the volume rendering loss in NeRF~\citep{mildenhall2020nerf}.




\subsection{Network Structure of LeC\texorpdfstring{$^2$}{2}O-NeRF}

\par\noindent\textbf{Occupancy network.}
The occupancy network $O$ in our LeC$^2$O-NeRF serves as a classification network to dispatch 3D points into different sub-networks, as depicted in Fig.~\ref{fig:framework}. $O$ predicts a vector of $n+1$ normalized occupancy values $O(\textbf{x})$ for a 3D point $\textbf{x}$. The first $n$ occupancy values correspond to the $n$ scene sub-networks in $\mathcal{S}$. The last occupancy value corresponds to the empty space network $E_e$. We use Top-$1$ operation to obtain the index $k$ of the Top-$1$ value in $O(\textbf{x})$. Then, we dispatch $\textbf{x}$ into the sub-network of index $k$. The occupancy value is multiplied by the output of the sub-network. 
This allows gradients from the main rendering loss to be propagated backward through the occupancy network, enabling the occupancy network to be trained together with the entire network. 

If $\textbf{x}$ is assigned to $E_e$, it implies that $\textbf{x}$ is unoccupied. It goes through $E_e$ and $H_e$ to predict density $\sigma$ and color $\textbf{c}$. If the assigned sub-network is one of the scene sub-networks, this indicates that $\textbf{x}$ is occupied. Then, $\textbf{x}$ goes through the corresponding scene sub-network and the head $H_s$ to predict $\sigma$ and $\textbf{c}$. After training the entire network, the occupancy of a 3D scene is encoded into the compact occupancy network $O$. Then, we can use $O$ as an occupancy predictor. An input point is unoccupied if the occupancy network dispatches it to $E_e$. In our implementation, the occupancy network contains 4 linear layers and a layer-norm layer.

\noindent \textbf{Sub-networks and heads.}
The proposed LeC$^2$O-NeRF is imbalanced because the sub-networks are different.
The scene network $\mathcal{S} = \{S_i, i=1...n\}$ contains $n$ sub-networks with the same architecture, each of which consists of 7 linear layers. They encode the occupied points. The prediction head $H_s$ for $\mathcal{S}$ is shared. $H_s$ also accepts the view direction $\textbf{d}$ and appearance embedding $\mbox{AE}$~\citep{martin2021nerf} as inputs to encode a view-dependent color. We use $n$ sub-networks in the scene network in order to enlarge its network capacity for encoding large-scale scenes.

The empty space network $E_e$ is defined as a \emph{tiny} network to encode unoccupied (\ie~empty space) points. We use an identity layer to directly feed-forward the input into the prediction head $H_e$. The tiny $E_e$ results in fewer parameters for empty space. As a result, $E_e$ tends to predict smooth density and color values and therefore favors the unoccupied points whose density is small and smooth. The scene network $\mathcal{S}$ is designed to contain much more network parameters than $E_e$, because occupied points contain significantly more important information.

\begin{figure}[t]
\begin{center}
   \includegraphics[width=\linewidth]{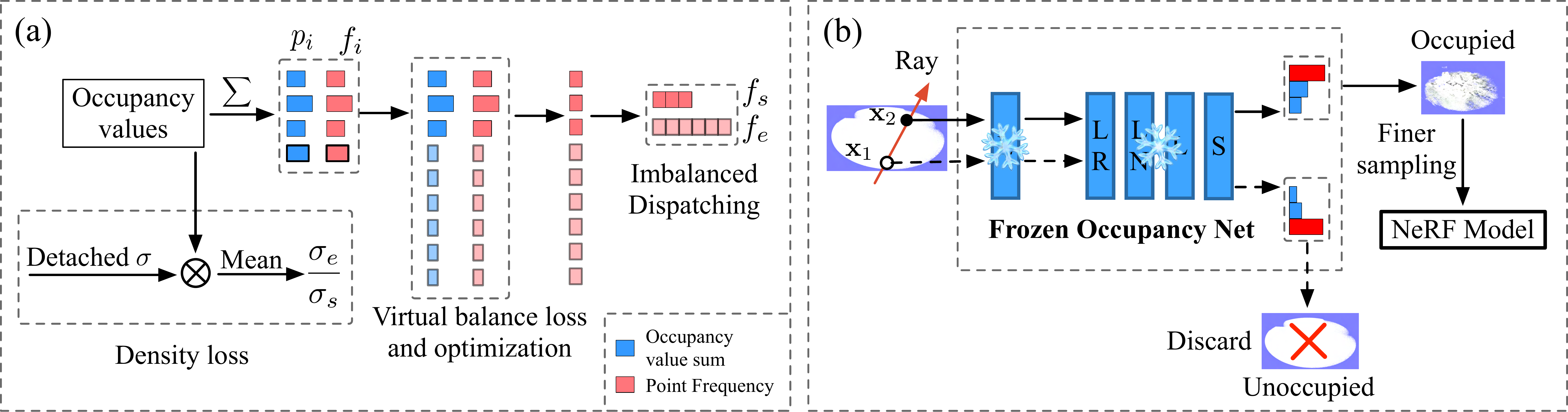}
\end{center}

   \caption{(a) The computation of the imbalanced occupancy loss and the density loss from occupancy values. (b) After training the occupancy network of a scene, we can use our frozen occupancy network to guide the sampling and training of NeRF methods.}
\label{fig:loss_sampling}
\end{figure}

\subsection{Large-Scale Occupancy Optimization Losses}
The occupancy network and imbalanced network structure cannot naturally learn reasonable occupancy without priors of the 3D scene. We further propose an imbalanced occupancy loss and a density loss to regularize the occupancy network to learn accurate occupancy for a large-scale scene, as depicted in Fig.~\ref{fig:loss_sampling}(a).

\par\noindent \textbf{Imbalanced occupancy loss.} In a 3D scene, the occupied and unoccupied 3D points are naturally imbalanced. A large portion of the 3D scene points is unoccupied. Our empty space network $O$ should secure more 3D points to faithfully learn the imbalanced nature of the scene. To accomplish the imbalanced classification, we design an imbalanced occupancy loss $L_o$ to directly control the portions of occupied and unoccupied points during the training. This loss not only can dispatch more points into $E_e$, but also can keep the number of points roughly the same for each scene sub-network. This means that $L_o$ is imbalanced for the empty space network while balanced for the scene sub-networks.

Our imbalanced occupancy loss is inspired by the balanced loss $L_b$ in~\citep{DBLP:conf/iclr/LepikhinLXCFHKS21}. We first introduce this balanced loss. It aims to dispatch a similar number of points to each sub-network. Let $n$ be the number of total sub-networks, and $f_i$ be the fraction of points dispatched into sub-network $i$. Then, $\sum{f_i}^2$ is minimized if all $f_i$ are equal. However, $\sum{f_i}^2$ is not differentiable, so it cannot be used as a loss function. As shown in~\citep{DBLP:conf/iclr/LepikhinLXCFHKS21}, it replaces one $f_i$ by a soft version $p_i$, where $p_i$ is the fraction of the occupancy values dispatched to sub-network $i$. Therefore, the balanced loss can be defined as $L_b = n\sum_{i=1}^n f_ip_i$. Under the optimal balance dispatching, $L_b$ will be $1$. Inspired by $L_b$, we define the imbalanced occupancy loss $L_o$. We can consider the empty space network $E_e$ as $v$ virtual sub-networks. The fraction of each virtual sub-network is thus $f_e/v$, and the fraction of the occupancy values is $p_e / v$. Then, we can compute the balanced loss for the $v$ virtual sub-network and the $n$ scene sub-networks. When $n+v$ sub-networks are balanced, $E_e$ can obtain more points. Hence, we define $L_o$ as:
\begin{equation}
    L_o = (n + v)\left( v \frac{f_e}{v}\frac{p_e}{v} + \sum_{i=1}^n f_ip_i\right) = (n + v)\left( \frac{f_e p_e}{v} + \sum_{i=1}^n f_ip_i\right)
\end{equation}
When optimal dispatching is achieved, $E_e$ obtains a portion of $v/(n+v)$ points. Each scene sub-network obtains a portion of $1/(n+v)$. $L_o$ is $1.0$. Therefore, $L_o$ can approximately control the ratio of occupied and unoccupied points. 
We typically set $n=8$, $v=80$ in our experiments. These values make the occupancy network dispatch about 85\% points to the empty space network.

\noindent \textbf{Density loss.} We design a density loss to explicitly guide the occupancy network $O$ to learn better occupancy. Our main idea is that the average density of the points dispatched to the empty space sub-network $E_e$ should be much smaller than that of the scene network $\mathcal{S}$. Let the set of points dispatched to $E_e$ and $\mathcal{S}$ be $\mathcal{X}$ and $\mathcal{Y}$, respectively. The average density $\sigma_e$ of $E_e$ is $\sigma_e = \frac{1}{|\mathcal{X}|} \sum_{i \in \mathcal{X}} \sigma_i $. The average density $\sigma_s$ for $\mathcal{S}$ is $\sigma_s = \frac{1}{|\mathcal{Y}|} \sum_{i \in \mathcal{Y}} \sigma_i $. Then, the ratio $\sigma_e / \sigma_s$ should be small if the occupancy is learned correctly. The problem is that the $\sigma_e / \sigma_s$ cannot affect the occupancy network. Therefore, we include the occupancy values in the computation of the mean density. The $\sigma_e$ and $\sigma_s$ can be rewritten as $\sigma_e = \frac{1}{|\mathcal{X}|} \sum_{i \in \mathcal{X}} o_i\sigma_i$ and $\sigma_s = \frac{1}{|\mathcal{Y}|} \sum_{i \in \mathcal{Y}} o_i\sigma_i $. The value $o_i$ used for a point of the scene network is the sum of the occupancy values for the $n$ scene sub-networks. Therefore, the density loss $L_d$ can be defined as:
\begin{equation}
    L_d = \frac{\sigma_e}{\sigma_s} = \frac{|\mathcal{Y}|}{|\mathcal{X}|} \frac{\sum_{i \in \mathcal{X}} o_i\sigma_i}{\sum_{i \in \mathcal{Y}} o_i\sigma_i}
\end{equation}
We detach $\sigma$ when computing $L_d$.
When $L_d$ is large, it optimizes the output of the occupancy network to make it dispatch correctly.

\par\noindent \textbf{Rendering loss.} Our network learns the occupancy during the training of NeRF. Therefore, our main optimization loss is the rendering loss~\citep{mildenhall2020nerf}. We sample $N$ 3D points along a ray $\textbf{r}$ and predict the density $\sigma_i$ and color $\textbf{c}_i$ for each 3D point $\textbf{x}_i$ by the network. We use $\sigma_i$ to compute $\alpha_i = 1-\mbox{exp}(-\sigma_i\delta_i)$, where $\delta_i$ is the distance of two nearby points. Then, we compute the transmittance $T_i=\mbox{exp}(-\sum_{j=1}^{i-1}\sigma_j\delta_j)$ of $\textbf{x}_i$ along the ray. The predicted color $\hat{C}(\textbf{r})$ is computed as $\hat{C}(\textbf{r}) = \sum_{i=1}^NT_i\alpha_i\textbf{c}_i$.
The rendering loss $L_r$ is computed by $\hat{C}(\textbf{r})$ and the ground-truth color $C(\textbf{r})$. Let the set of rays be $\mathcal{R}$. $L_r$ is defined as $L_r = \sum_{r \in \mathcal{R}}\left\|\hat{C}(\textbf{r}) - C(\textbf{r})\right\|_2^2$.

\noindent \textbf{Final loss.} The final loss $L_f$ is the weighted sum of $L_r$, $L_o$ and $L_d$. $L_f = w_r L_r + w_o L_o + w_d L_d$, where $w_r$, $w_o$, and $w_d$ are corresponding loss weights.

\subsection{Large-Scale Occupancy as Guidance}
When training the occupancy network, unoccupied points still need gradients for optimization, which consumes memory and computation. Since our occupancy network converges very fast, we can freeze it after it converges and use it to filter unoccupied points for a NeRF network, as shown in Fig.~\ref{fig:loss_sampling}(b). $O$ is much smaller than the main NeRF network, so the training can be significantly accelerated.  

In the guided training, we first sample a set of coarse samples and send them into the frozen occupancy network $O$ to discard unoccupied points, typically 85\% points from our observation. Then, we split the reserved samples to obtain finer samples. These two steps can reduce the number of points sent into $O$. In the experiments, we typically sample 128 samples along a ray and use the occupancy to filter the samples and split each occupied sample into 8 new samples.


\section{Experiments}

\vspace{-5pt}
\subsection{Datasets}
We use two publicly available large-scale datasets for evaluation. The Mega-NeRF dataset~\citep{Turki_2022_CVPR} consists of the Building, Rubble, Residence, Sci-Art, and Campus scenes. Each of them contains from $2k$ to $6k$ images with a resolution of about $5k\times 3k$. The Block-NeRF dataset~\citep{Tancik_2022_CVPR} contains a scene with $12k$ images with a resolution of about $1k \times 1k$. 

\vspace{-5pt}
\subsection{Metrics and Visualization}\label{sec:Metricsandvisualization}

We evaluate the occupancy accuracy with Occupancy Metrics and apply the occupancy on the sampling of state-of-the-art NeRF methods~\citep{mueller2022instant, mi2023switchnerf} to compute the Image Reconstruction Metrics.

\noindent \textbf{Occupancy metrics.} 
We evaluate the occupancy classification accuracy. The ground-truth occupancy is usually not available in real-world large-scale NeRF datasets. As a fully-trained NeRF without using occupancy can obtain a good estimation of the geometry of the scene, we use it as a good reference for evaluation. We extract depth maps predicted by the large-scale Switch-NeRF~\citep{mi2023switchnerf} and convert them into an occupancy grid. Then, we also convert our learned occupancy into another occupancy grid by sampling and evaluating point occupancy. The occupancy accuracy is computed by comparing the converted occupancy grids. 

\noindent \textbf{Image reconstruction metrics.} 
We use our learned occupancy to guide the training of several representative NeRF methods, including Instant-NGP (INGP) \citep{mueller2022instant} and the large-scale Switch-NeRF~\citep{mi2023switchnerf}. We use PSNR, SSIM~\citep{wang2004image} (both higher is better), and LPIPS~\citep{zhang2018unreasonable} (lower is better) to evaluate the validation images.

\noindent \textbf{Occupancy visualization.} We visualize the occupancy as point clouds. We sample and merge 3D points of rays in the validation images. These point clouds are visualized by two methods. The first one is to directly visualize the predicated color of each point. The second one uses the $\alpha = 1-\mbox{exp}(-\sigma_i \delta_i)$ as an additional channel to show the color and transparency of the point clouds. The unoccupied points should be largely transparent. The two visualization methods complement each other for better visualization of the occupancy.
\begin{table}[t]

  \centering
  \caption{Accuracy, Precision, Recall, F1-Score, parameter number, and occupancy ratio of different occupancy methods. Our method clearly outperforms the occupancy grid with more compact parameter sizes and better occupancy ratios.}
  \resizebox{0.9\textwidth}{!}{
    \begin{tabular}{c|cc|cc|cc|cc|cc}
    \toprule
    Dataset & \multicolumn{2}{c|}{Sci-Art} & \multicolumn{2}{c|}{Campus} & \multicolumn{2}{c|}{Rubble} & \multicolumn{2}{c|}{Building} & \multicolumn{2}{c}{Residence} \\
    \midrule
    Method & Grid  & Ours  & Grid  & Ours  & Grid  & Ours  & Grid  & Ours  & Grid  & Ours \\
    \midrule
    Accuracy & \textbf{0.912} & 0.904 & 0.619 & \textbf{0.684} & 0.697 & \textbf{0.712} & 0.521 & \textbf{0.711} & 0.656 & \textbf{0.703} \\
    Precision & 0.315 & \textbf{0.339} & 0.371 & \textbf{0.437} & 0.269 & \textbf{0.319} & 0.183 & \textbf{0.322} & 0.232 & \textbf{0.314} \\
    Recall & 0.519 & \textbf{0.795} & 0.746 & \textbf{0.883} & 0.666 & \textbf{0.914} & 0.549 & \textbf{0.683} & 0.634 & \textbf{0.958} \\
    F1-Score & 0.392 & \textbf{0.476} & 0.496 & \textbf{0.584} & 0.383 & \textbf{0.473} & 0.274 & \textbf{0.438} & 0.339 & \textbf{0.473} \\
    \midrule
    Para. Number & 2.0M  & \textbf{0.15M} & 2.0M  & \textbf{0.15M} & 2.0M  & \textbf{0.15M} & 2.0M  & \textbf{0.15M} & 2.0M  & \textbf{0.15M} \\
    Occupancy ratio & 22.8\% & \textbf{13.0\%} & 34.0\% & \textbf{14.5\%} & 33.6\% & \textbf{13.0\%} & 44.0\% & \textbf{15.0\%} & 37.5\% & \textbf{15.9\%} \\
    \bottomrule
    \end{tabular}%
    }
  \label{tab:occupancy-metric}%
\end{table}%

\subsection{Implementation Details}
When training the occupancy network on the Mega-NeRF dataset, we use 8 sub-networks for the scene network. The occupancy network contains one input layer, two inner layers, one layer-norm layer, and one output layer. The channel number of the main layers is set as 256. The empty space network is an identity layer. We set $w_r = 1.0$, $w_o = 0.0005$, $w_d = 0.1$ and $v=80$. We sample 512 points for each ray. We train the occupancy for $40k$ steps. The training of the occupancy network takes from 1.6h to 1.8h. The learning rate is set as $5 \times 10^{-4}$.
When training the occupancy network on the Block-NeRF dataset, we use Mip embedding proposed in~\citep{barron2021mipnerf}. 
$w_d$ is set as $0.005$ and $v$ is set as 40.

When applying our learned occupancy network on NeRF methods such as the Instant-NGP (INGP)~\citep{mueller2022instant} and Switch-NeRF~\citep{mi2023switchnerf}, we use the occupancy network to guide their sampling. To compare with the occupancy grid, we employ the OccGridEstimator from NeRFAcc~\citep{li2022nerfacc}. The grid size is set as the default (i.e., $128^3$). The OccGridEstimator is updated with the main network. The main results are obtained by training on 2 NVIDIA RTX 3090 GPUs for INGP and 8 NVIDIA RTX 3090 GPUs for Switch-NeRF. We sample 8192 rays for the Mega-NeRF dataset and 13312 rays for the Block-NeRF dataset. We align the training time of our methods with the grid methods trained with 500k iterations.

\subsection{Benchmark Performance}

 \begin{wraptable}[16]{r}{0.515\textwidth}
  \centering
  \caption{The image accuracy on Block-NeRF dataset~\citep{Tancik_2022_CVPR}. INGP+Ours outperforms INGP~\citep{mueller2022instant} by a PSNR of \textbf{2.36}. Switch+Ours not only outperforms Switch-NeRF~\citep{mi2023switchnerf}, but also outperforms Switch+Grid by a PSNR of \textbf{0.84}. INGP-based methods are trained with 20.0h. Switch-NeRF-based methods are trained with 24.0h.}
  \resizebox{0.8\linewidth}{!}{
    \begin{tabular}{c|cccc}
    \toprule
    Method & PSNR$\uparrow$  & SSIM$\uparrow$  & LPIPS$\downarrow$ \\
    \midrule
    INGP & 19.49 & 0.701 & 0.558 \\
    INGP+Ours & \textbf{21.85} & \textbf{0.738} & \textbf{0.507} \\
    \midrule
    Switch	& 22.85 & 0.742 & 0.515 \\
    Switch+Grid & 22.26 & 0.740 & 0.511 \\
    Switch+Ours & \textbf{23.10} & \textbf{0.751} & \textbf{0.498} \\
    \bottomrule
    \end{tabular}%
  \label{tab:main-block-results}%
  }

\end{wraptable} 

\textbf{Occupancy Metrics.} We evaluate our occupancy accuracy with the Occupancy Metrics. Since the unoccupied and occupied points are highly imbalanced, we report the Accuracy, Precision, Recall, and F1-Score to complement each other. In Table~\ref{tab:occupancy-metric}, our learned occupancy can clearly outperform the Occupancy Grid in almost all the metrics on all Mega-NeRF scenes, with clearly more compact parameter sizes. Notably, our network is much better on Recall, indicating that it is good at correctly predicting the occupied points, which is critical for better NeRF optimization. The occupancy ratio in Table~\ref{tab:occupancy-metric} means the ratio of occupied points to go through the main NeRF. Our occupancy network also retains fewer points than the occupancy grid while achieving better accuracy. These results demonstrate that our method can predict more accurate and compact occupancy.

\noindent \textbf{Image Reconstruction Metrics.} Table~\ref{tab:main-block-results} shows the results of applying our learned occupancy on Instant-NGP (INGP)~\citep{mueller2022instant} and Switch-NeRF~\citep{mi2023switchnerf} on Block-NeRF dataset. Our method significantly surpasses both Instant-NGP~\citep{mueller2022instant} (INGP) and Switch-NeRF with an occupancy grid (Switch+Grid) in terms of PSNR, with margins of \textbf{2.36} and \textbf{0.84}, respectively. Given the substantial size of Block-NeRF dataset, which comprises $12k$ images, the results highlight the superiority of our method when compared to the occupancy grid method. Note that the training time of our method includes our occupancy training time for fair comparisons.

\begin{table}[htbp]
  \centering
  \caption{The accuracy and training time on Mega-NeRF dataset~\citep{Turki_2022_CVPR}. Our method on Instant-NGP (NGP+Ours) clearly outperforms the occupancy grid INGP~\citep{mueller2022instant}. Our method on Switch-NeRF (Switch+Ours) clearly outperform Switch+Grid and Switch-NeRF (Switch)~\citep{mi2023switchnerf}. The occupancy grid cannot successfully speed up the training without decreasing the accuracy. Switch-NeRF-based methods are trained by 13.6h. NGP-based methods are trained by 11.6h.}
  \resizebox{\textwidth}{!}{
    \begin{tabular}{c|ccc|ccc|ccc|ccc|ccc}
  \toprule
    Dataset & \multicolumn{3}{c|}{Sci-Art} & \multicolumn{3}{c|}{Campus} & \multicolumn{3}{c|}{Rubble} & \multicolumn{3}{c|}{Building} & \multicolumn{3}{c}{Residence} \\
    \midrule
    Metric  & PSNR$\uparrow$  & SSIM$\uparrow$  & LPIPS$\downarrow$ & PSNR$\uparrow$  & SSIM$\uparrow$  & LPIPS$\downarrow$ & PSNR$\uparrow$  & SSIM$\uparrow$  & LPIPS$\downarrow$ & PSNR$\uparrow$  & SSIM$\uparrow$  & LPIPS$\downarrow$ & PSNR$\uparrow$  & SSIM$\uparrow$  & LPIPS$\downarrow$ \\
    \midrule
    Switch & 25.46 & 0.762 & 0.400 & 22.76 & 0.507 & 0.659 & 23.58 & 0.519 & 0.546 & 20.50 & 0.517 & 0.526 & 21.77 & 0.611 & 0.501  \\
    Switch+Grid & 25.48 & 0.760 & 0.413 & 22.75 & 0.500 & 0.671 & 23.69 & 0.522 & 0.549 & 20.33 & 0.495 & 0.547 & \textbf{22.18} & 0.622 & 0.500 \\
    Switch+Ours & \textbf{26.04} & \textbf{0.772} & \textbf{0.398} & \textbf{23.21} & \textbf{0.517} & \textbf{0.635} & \textbf{23.96} & \textbf{0.548} & \textbf{0.516} & \textbf{20.64} & \textbf{0.522} & \textbf{0.517} & 22.10 & \textbf{0.626} & \textbf{0.485} \\
    \midrule
    NGP   & 23.98 & 0.724 & 0.445 & 21.76 & 0.475 & 0.677 & 22.94 & 0.498 & 0.572 & 19.48 & 0.454 & 0.585 & 21.27 & 0.591 & 0.515  \\
    NGP+Ours & \textbf{24.50} & \textbf{0.754} & \textbf{0.412} & \textbf{22.83} & \textbf{0.518} & \textbf{0.623} & \textbf{23.66} & \textbf{0.558} & \textbf{0.501} & \textbf{20.33} & \textbf{0.511} & \textbf{0.518} & \textbf{21.77} & \textbf{0.626} & \textbf{0.473} \\
    \bottomrule
    \end{tabular}%
    }
  \label{tab:main-mega-results}%
\end{table}%

Table~\ref{tab:main-mega-results} shows the results of Switch-NeRF (Switch)~\citep{mi2023switchnerf}, Switch-NeRF with an occupancy grid (Switch+Grid), Switch-NeRF with our learned occupancy network (Switch+Ours), Instant-NGP (INGP)~\citep{mueller2022instant}, and Instan-NGP with our learned occupancy netowrk (INGP+Ours), on the Mega-NeRF dataset~\citep{Turki_2022_CVPR}. Instant-NGP~\citep{mueller2022instant} already incorporates an occupancy grid to guide the training. We align the training time to the grid-based methods.
Note that we include the occupancy training time in Switch+Ours and NGP+Ours for a fair comparison. 

\begin{figure}[t]

\begin{center}
   \includegraphics[width=\linewidth]{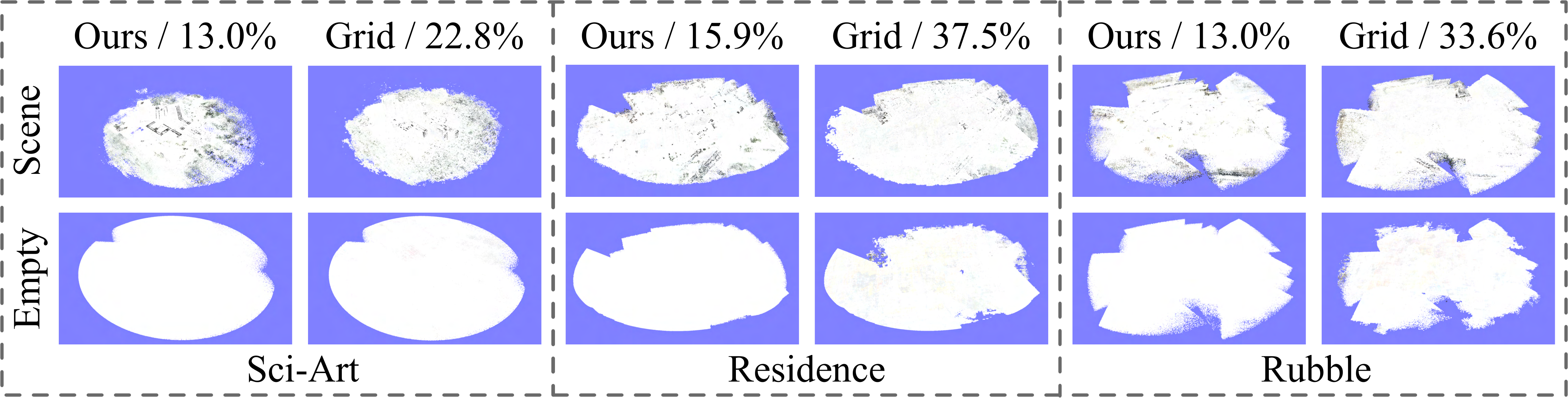}
\end{center}

   \caption{The visualization of our occupancy and the grid occupancy as point clouds. Our predicted occupied points (scene surface points) are cleaner and have fewer points than the grid occupancy. They fit the surface of the buildings more compactly.}

\label{fig:ours_grid_points}
\end{figure}

We highlight the best values among Switch, Switch+Grid, and Switch+Ours, and the best values between NGP and NGP+Ours.
As shown in Table~\ref{tab:main-mega-results}, our method consistently outperforms Switch, NGP, and Switch+Grid. Therefore, our method is significant to speed up the training of Switch-NeRF and NGP while achieving superior accuracy. Notably, the Switch+Grid does not obtain better results than Switch-NeRF. This means that on a challenging large-scale scene, the occupancy grid cannot successfully speed up the training without sacrificing accuracy. In contrast, our occupancy network can largely improve the accuracy. We visualize the point clouds of occupancy in Fig.~\ref{fig:ours_grid_points}. The point clouds show that our network can learn more compact and clean occupancy than the occupancy grid. We also provide the visualization comparison of rendered images in Fig.~\ref{fig:imagecompare} and a video in the supplementary files.

\begin{figure}[htbp]

\begin{center}
   \includegraphics[width=\linewidth]{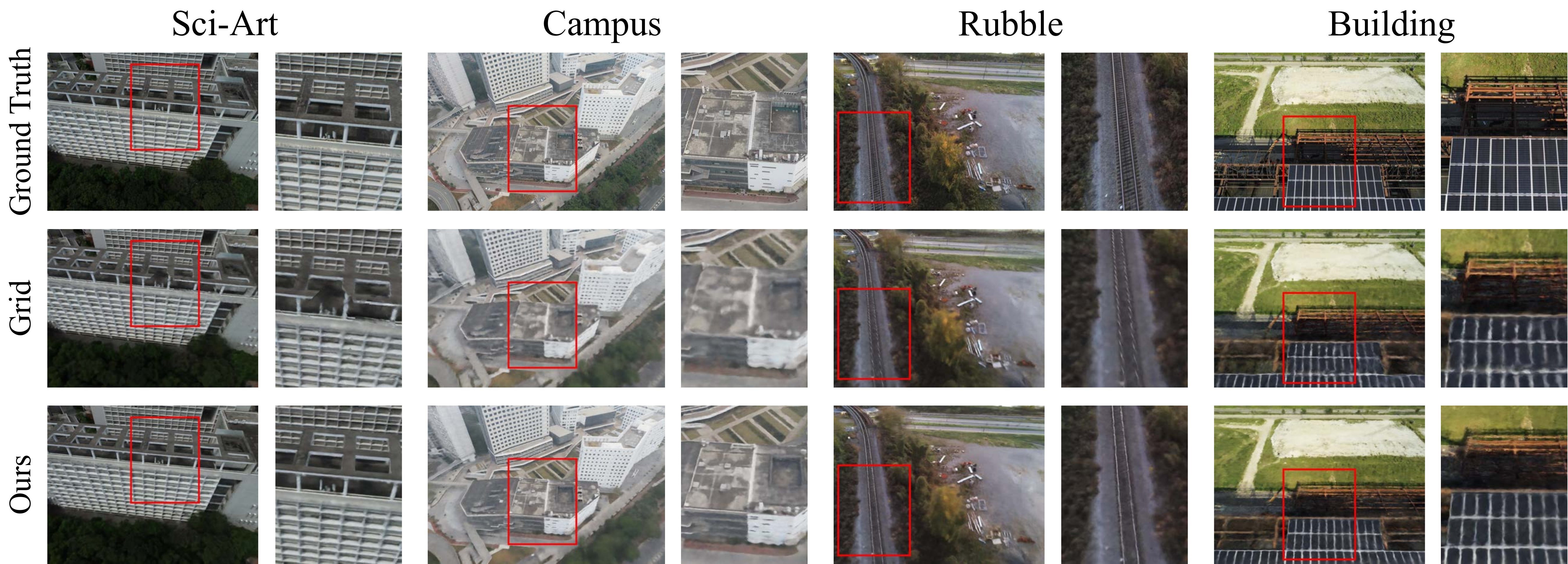}
\end{center}

   \caption{The rendered images of our occupancy network and the occupancy grid based on Switch-NeRF. Our method can obtain more complete, clean, and high-quality images.
}

\label{fig:imagecompare}
\end{figure}

\subsection{Ablation Study}
In this section, we perform several ablations to analyze the designs of our imbalanced network structure, the density loss, and the learned occupancy. The experiments are performed by applying our occupancy on Switch-NeRF and the Sci-Art scene for 40K occupancy steps and 500K NeRF steps if not specified.

\begin{table}[htbp]
\parbox{.48\linewidth}{
\centering
\caption{Ablation on the structure of $E_e$ \textbf{with} occupancy loss $L_o$ \textbf{without} the density loss $L_d$. 7-layer, 4-layer, and Identity mean using a 7-layer MLP, a 4-layer MLP, or an Identity layer in the empty space network. Larger empty space networks cannot learn reasonable occupancy, while our imbalanced structure with an Identity layer can learn good occupancy.
}
\resizebox{0.8\linewidth}{!}{
\begin{tabular}{c|ccc}
    \toprule
    Method & PSNR$\uparrow$  & SSIM$\uparrow$  & LPIPS$\downarrow$ \\
    \midrule
    7-layer & 20.23 & 0.631 & 0.506 \\
    4-layer & 23.62 & 0.701 & 0.463 \\
    Identity & \textbf{26.30} & \textbf{0.781} & \textbf{0.383} \\
    \bottomrule
\end{tabular}
}

\label{tab:ablations_heter}
}
\hfill
\parbox{.48\linewidth}{
\centering
\caption{Ablation study on the memory usage and accuracy of our method and the occupancy grids with different resolutions. With a resolution of $256^3$ (Grid-256), the occupancy grid method slows down dramatically and obtains much worse accuracy than ours trained with the same time. Moreover, Grid-256 consumes about $4.5\times$ memory than our method. All methods are trained with 14.1h.}
\resizebox{0.9\linewidth}{!}{
\begin{tabular}{c|ccccc}
    \toprule
    Method & PSNR$\uparrow$  & SSIM$\uparrow$  & LPIPS$\downarrow$ & Mem.$\downarrow$ \\
    \midrule
    Switch-NeRF & 25.46 & 0.762 & 0.400 & 10.5G \\
    Grid-128 & 25.48 & 0.760 & 0.413 & 5.8G \\
    Grid-256 & 24.75 & 0.728 & 0.448 & 12.5G \\
    Ours  & \textbf{26.04} & \textbf{0.772} & \textbf{0.398} & \textbf{2.7G} \\
    \bottomrule
\end{tabular}}
\label{tab:gridresolutionresult}
}
\end{table}

\noindent \textbf{Imbalanced network structure.}
We perform experiments to show that our designed imbalanced network structure with the imbalanced occupancy loss can learn the occupancy. We set the empty space network $E_e$ to different network sizes. 7-layer means using a 7-layer MLP in $E_e$, the same as the scene sub-networks, creating a balanced structure. 4-layer uses a smaller 4-layer MLP in $E_e$. Identity means that we use an identity layer in $E_e$, which is our proposed imbalanced network structure. 

\begin{figure}[t]
\begin{center}
   \includegraphics[width=\linewidth]{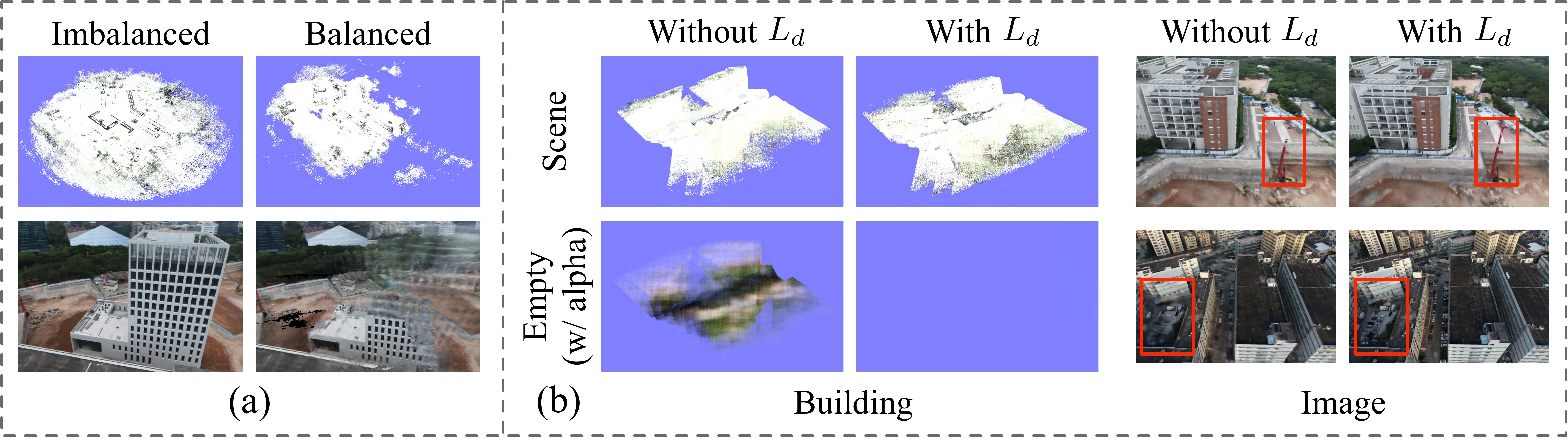}
\end{center}
   \caption{(a) The occupied points and images with the imbalanced and balanced networks. Our imbalanced network learns complete occupancy and images. The balanced structure cannot distinguish the occupied and unoccupied regions. (b) Point clouds of the scene network and the empty space network with and without the density loss. We visualize the point clouds of the empty space network with transparency related to alpha values (see Sec.~\ref{sec:Metricsandvisualization}) to better show whether the points are empty or not. With $L_d$, our imbalanced structure can learn better occupancy and thus the points of $E_e$ are all transparent. With $L_d$, the images are more complete in challenging regions.}
\label{fig:density_loss}
\end{figure}

As shown in Table~\ref{tab:ablations_heter}, the rendering accuracy of the 7-layer balanced structure largely dropped compared to our imbalanced empty space network with an Identity layer. Our imbalanced structure can obtain reasonable accuracy. Note that the imbalanced occupancy loss is used while the density loss is \textbf{not} used in these experiments. As shown in Fig.~\ref{fig:density_loss}(a), the scene network of our imbalanced structure handles the full occupied points. The scene network of the balanced structure only handles a part of the occupied points. This means that the balanced structure cannot learn reasonable occupancy, and its rendered image is thus of low quality. These experiments show that, to implicitly model the imbalanced occupancy of a 3D scene, it is important to design an imbalanced network. 






\begin{wraptable}[10]{r}{0.5\textwidth}
  \centering
  \caption{Ablation on the density loss $L_d$ on the Building scene. $L_d$ can help our occupancy network learn better occupancy and achieve better accuracy.}
  \resizebox{0.8\linewidth}{!}{

    \begin{tabular}{c|ccc}
    \toprule
    Method & PSNR$\uparrow$  & SSIM$\uparrow$  & LPIPS$\downarrow$ \\
    \midrule
    Ours w/o $l_d$ & 20.59 & 0.516 & 0.521 \\
    Ours & \textbf{20.79} & \textbf{0.531} & \textbf{0.508} \\
    \bottomrule
    \end{tabular}%
    }
  \label{tab:ablations_density}%
\end{wraptable} 
\noindent \textbf{Density loss.} We ablate on the density loss $L_d$ on the Building scene in Table~\ref{tab:ablations_density}. Our full method achieves better accuracy than that without the density loss. As shown in Fig.~\ref{fig:density_loss}(b), with $L_d$, the network can separate the occupied and unoccupied points more clearly, and the rendered images contain fewer artifacts in the challenging regions. These experiments show that, our density loss can provide more explicit information to the occupancy network and make the occupancy network learn more accurate occupancy.

\noindent \textbf{Memory usage.} We analyze the memory usage and the accuracy of our method, and the occupancy grids with different resolutions to better demonstrate the advantages of our compact occupancy network. 
As shown in Table~\ref{tab:gridresolutionresult}, our method has the lowest memory usage compared to Switch-NeRF~\citep{mi2023switchnerf} and occupancy grids with resolutions of $128^3$ (Grid-128) and $256^3$ (Grid-256). Notably, Switch-NeRF~\citep{mi2023switchnerf} and Grid-256 consume about $4.5\times$ more memory than ours while still achieving inferior results. The Grid-256 slows down dramatically than Grid-128 and obtains worse results when trained with the same time. This study clearly shows the advantage of the compactness of our proposed occupancy network.


\begin{figure}[t]
\begin{center}
   \includegraphics[width=\linewidth]{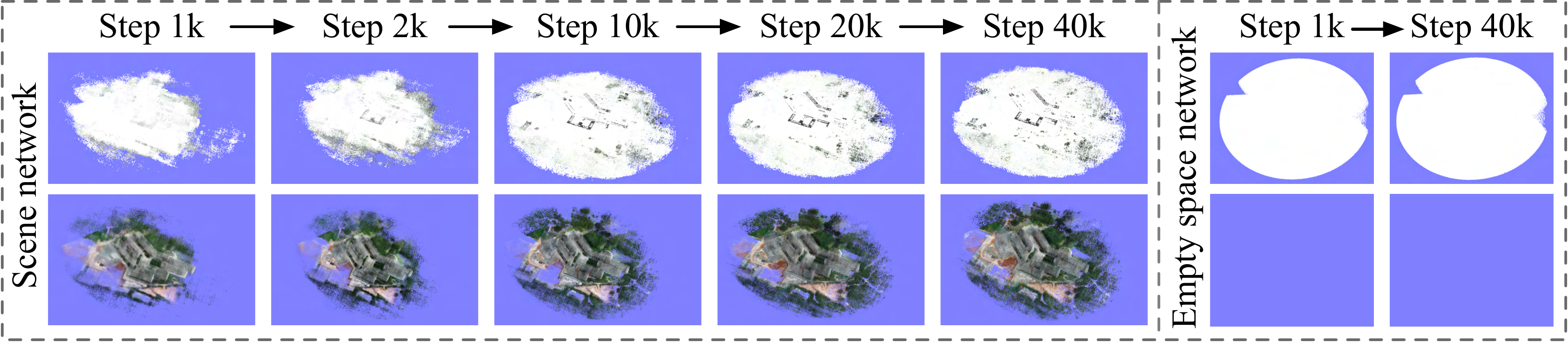}
\end{center}

   \caption{The point clouds dispatched to the scene network and the empty space network at each step. The scene network converges fast to the whole occupied area. The points in the empty space network consistently have very small opacity, resulting empty figures for the point cloud of the empty space network. Our network can learn accurate occupancy with only $20k$ to $40k$ steps. The two rows visualize the point clouds without and with transparency respectively as described in Sec.~\ref{sec:Metricsandvisualization}.}
\label{fig:occupancysteppoints}
\end{figure}

\begin{figure}[t]
	\centering
	\begin{subfigure}[b]{0.24\linewidth}
		\centering
		\includegraphics[width=\linewidth]{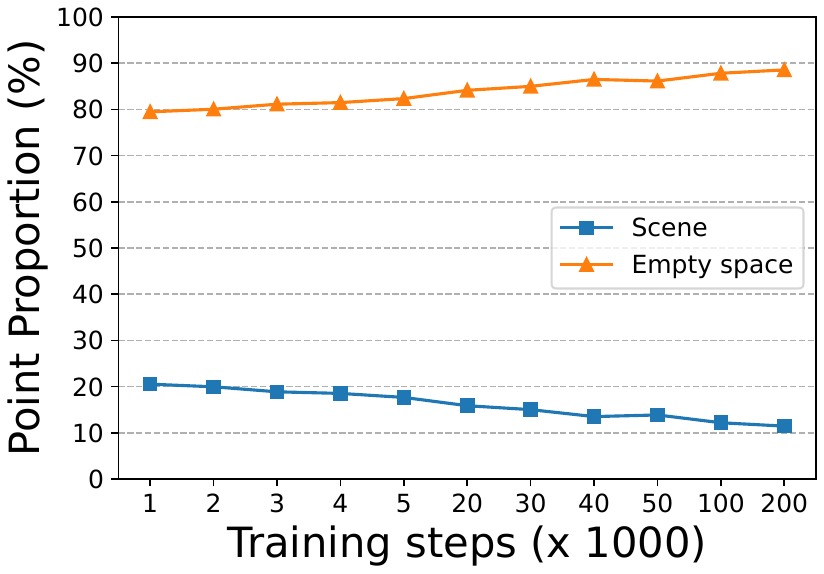}
		\caption[]%
		{{Point proportions.}}    
		\label{fig:pointproportions}
	\end{subfigure}
	\begin{subfigure}[b]{0.24\linewidth}   
		\centering 
		\includegraphics[width=\linewidth]{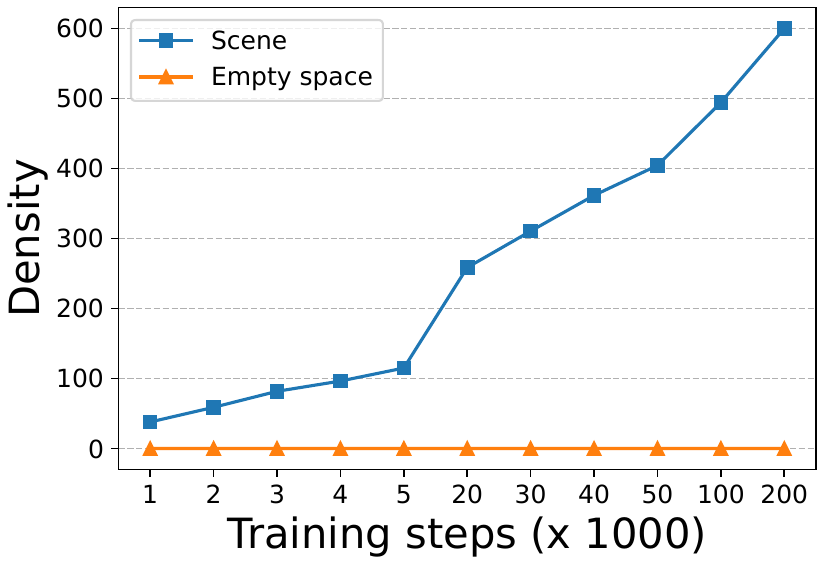}
		\caption[]%
		{{Mean density.}}    
		\label{fig:densityvalues}
	\end{subfigure}
	\begin{subfigure}[b]{0.24\linewidth}   
		\centering 
		\includegraphics[width=\linewidth]{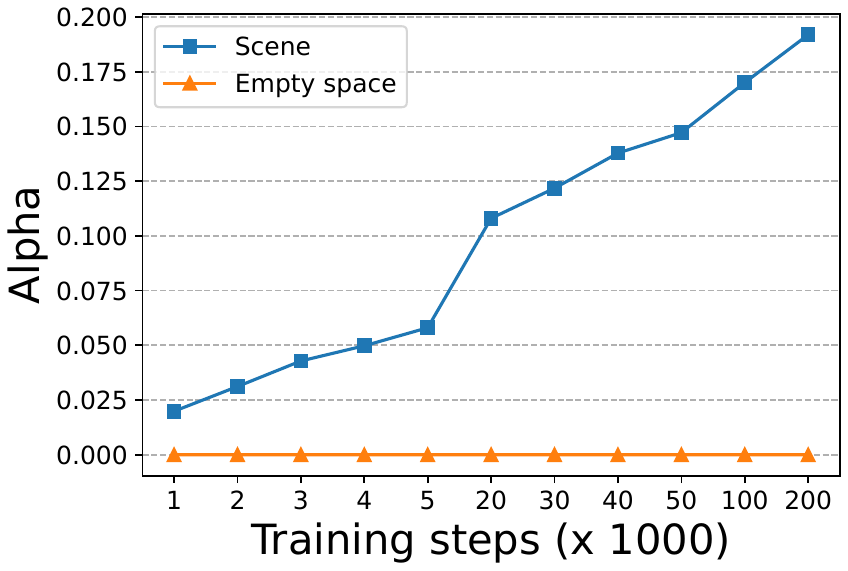}
		\caption[]%
		{{Mean alpha.}}    
		\label{fig:alphavalues}
	\end{subfigure}
	\begin{subfigure}[b]{0.24\linewidth}  
		\centering 
		\includegraphics[width=\linewidth]{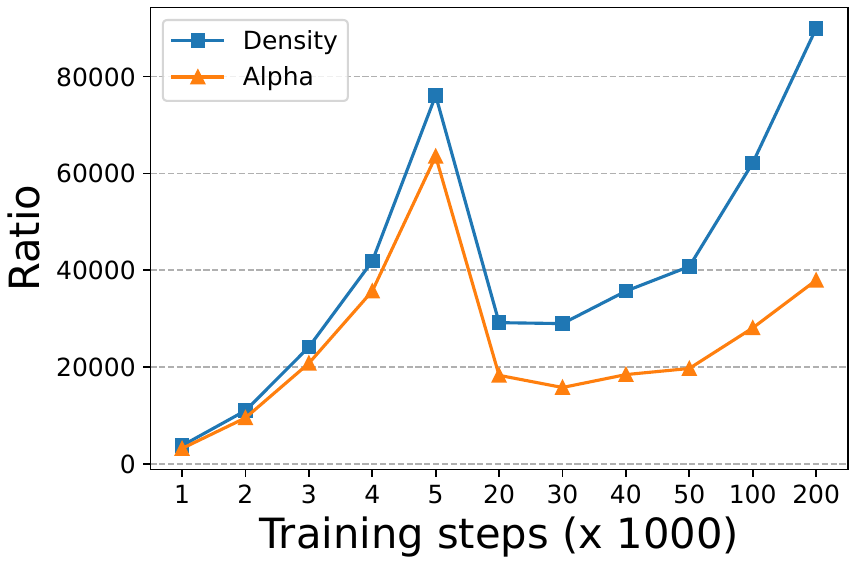}
		\caption[]%
		{{Ratio.}}    
		\label{fig:densityalpharatio}
	\end{subfigure}
	\caption[]
	{The statistics of the scene network $\mathcal{S}$ and the empty space network $E_e$ at different training steps. (a) The portion of the points in $\mathcal{S}$ and $E_e$. (b) (c) The mean density values and alpha values of the points in $\mathcal{S}$ and $E_e$. (d) The density value ratio and alpha value ratio between the points in $\mathcal{S}$ and $E_e$.} 
	\label{fig:occupancyanalysisfigures}
\end{figure}

\noindent \textbf{Occupancy analysis.} 
We analyze the occupancy statistics related to the points of the scene network and the empty space network with respect to the occupancy training steps on the evaluation images of the Sci-Art scene in Fig.~\ref{fig:occupancyanalysisfigures}.
Fig.~\ref{fig:pointproportions} is the portion of points in the scene network $\mathcal{S}$ and the empty space network $E_e$ of different training steps. There are consistently more than 80\% points in $E_e$. 
This figure verifies the effectiveness of our imbalanced occupancy loss. It also shows that we can speed up the training largely if we use the learned occupancy to guide the sampling of points.
Fig.~\ref{fig:densityvalues} and Fig.~\ref{fig:alphavalues} show mean density values and alpha values of the points in $\mathcal{S}$ and $E_e$. The points of $\mathcal{S}$ have clearly much larger densities and alpha values than those of $E_e$. The values of points in $E_e$ are nearly zero. This indicates that our network can effectively dispatch points according to their densities.
Fig.~\ref{fig:densityalpharatio} shows the density value ratio and the alpha value ratio between points in $\mathcal{S}$ and $E_e$. The values of points in $\mathcal{S}$ are several magnitudes larger than those in $E_e$. As the ratios are the direct target of the density loss, they can fully validate the effectiveness of our designed density loss.

\begin{figure}[htbp]
\begin{center}
   \includegraphics[width=0.9\linewidth]{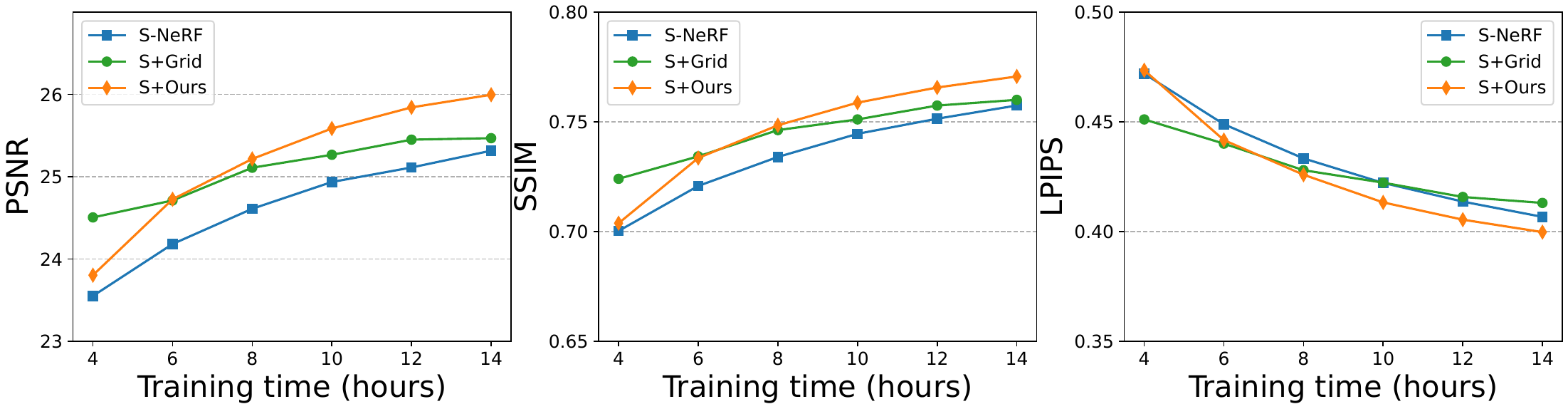}
\end{center}
   \caption{Analysis of training time and accuracy: our method (S+Ours) demonstrates remarkable convergence speed when compared to the grid-based occupancy (S+Grid) and the original Switch-NeRF (S-NeRF)~\citep{mi2023switchnerf} on Sci-Art. Note that the training time of our method includes the training time of our occupancy network.}

\label{fig:moe_occupancy_accuracy_time}
\end{figure}

Fig.~\ref{fig:occupancysteppoints} visualizes the point clouds dispatched to the scene network and the empty space network. The points in the scene network have larger opacity. They cover the whole scene surface quickly only after $10k$ steps of training. The points in the empty space network consistently present very small opacity, indicating that they are empty. The first row visualizes the points without transparency and the second row visualizes the points with transparency.

The extensive analysis of the occupancy clearly shows that the proposed LeC$^2$O-NeRF can learn the occupancy of a large-scale 3D scene accurately and quickly. It can be effectively encoded via our compact occupancy network.

\noindent \textbf{Accuracy of different training times.}
We analyze the detailed accuracy of our method on Sci-Art with respect to the training time in Fig.~\ref{fig:moe_occupancy_accuracy_time}. Our method on Switch-NeRF (S+Ours) demonstrates remarkable convergence speed compared to the grid-based occupancy on Switch-NeRF (S+Grid) and the original Switch-NeRF~\citep{mi2023switchnerf} (S-NeRF). Note that the training time of our method in this figure already includes the training time of our occupancy network.




    

\section{Conclusion}

In this paper, we have proposed LeC$^2$O-NeRF to learn continuous and compact occupancy for large-scale scenes. We achieve this by our core designs of a compact occupancy network, an imbalanced occupancy loss, a novel imbalanced network structure, and a density loss. Experiments on challenging large-scale datasets have shown that our learned occupancy clearly outperforms the occupancy grid and can achieve superior accuracy with much less time. Since occupancy is a very important concept in many 3D research areas, this work will offer more inspiration to the research of learning and representation of occupancy.

\bibliography{main}
\bibliographystyle{iclr2025_conference}


\end{document}

%% file: math_commands.tex
\usepackage{amsmath,amsfonts,bm}

\def\eqref#1{equation~\ref{#1}}

\def\1{\bm{1}}

\DeclareMathAlphabet{\mathsfit}{\encodingdefault}{\sfdefault}{m}{sl}
\SetMathAlphabet{\mathsfit}{bold}{\encodingdefault}{\sfdefault}{bx}{n}

